\documentclass[runningheads]{llncs}
\usepackage[T1]{fontenc}
\usepackage{graphicx,verbatim}
\usepackage{multirow}
\usepackage{amssymb}
\usepackage{amsmath} 
\usepackage{booktabs}
\usepackage{xcolor}
\usepackage[colorlinks,
linkcolor=red,
anchorcolor=blue,
citecolor=blue,
urlcolor=magenta]{hyperref}
\usepackage[capitalize]{cleveref}
\usepackage{pifont}
\usepackage[misc]{ifsym}

\begin{document}
%


\title{CardiacCLIP: Video-based CLIP Adaptation for LVEF Prediction in a Few-shot Manner}
\titlerunning{CardiacCLIP: Video-based CLIP for LVEF Prediction}

%
\author{Yao DU\inst{1} \and
Jiarong GUO\inst{1} \and
Xiaomeng LI\inst{1}\thanks{Corresponding Author}   }

\authorrunning{Y. DU et al.}
%
\institute{Department of Electronic and Computer Engineering, The Hong Kong University of Science and Technology, Hong Kong SAR, China \\
\email{eexmli@ust.hk}}



\maketitle             

\begin{abstract}

Echocardiography is a vital non-invasive modality for cardiac assessment, with left ventricular ejection fraction (LVEF) serving as a key indicator of heart function. Existing LVEF estimation methods depend on large-scale annotated video datasets, which are costly and limit adaptability across various clinical settings. Recent vision-language models for echocardiography, such as EchoCLIP, apply image-to-text pretraining but fail to capture crucial temporal dynamics and localized cardiac structures essential for accurate diagnosis. To address these challenges, we propose \textbf{CardiacCLIP}, a video-based framework that enhances LVEF prediction through attention-based frame aggregation and multi-resolution input scaling. Specifically, we introduce MFL~(Multi Frame Learning), a novel attention-based mechanism for selectively fusing informative frames, and EchoZoom, a multi-scale feature extraction strategy that refines spatial representations of cardiac structures. As a novel adaptation of CLIP models for few-shot echocardiogram video analysis, our approach significantly improves diagnostic accuracy, reducing MAE by 2.07 on the EchoNet-Dynamic dataset under 1-shot setting.
The code is available at {\tt\small \url{https://github.com/xmed-lab/CardiacCLIP}}.


\end{abstract}

\section{Introduction}
\label{sec:intro}

Left ventricular ejection fraction (LVEF) is a fundamental measure of cardiac function, widely used for diagnosing and monitoring cardiac conditions such as heart failure and cardiomyopathy~\cite{douglas2011accf,hughes2021deep,papolos2016us}. 
Echocardiography, as a non-invasive and cost-effective imaging modality, is the primary tool for assessing LVEF in clinical practice~\cite{ouyang2019echonet,yang2023graphecho,yang2024cardiacnet}. However, estimating LVEF remains a challenging task due to its dependence on expert interpretation, inter-operator variability, and the complex temporal dynamics of cardiac motion~\cite{pellikka2018variability,yang2024bidirectional}. 
Manual assessment is time-consuming and prone to subjectivity, highlighting the need for automated solutions that can improve efficiency and accuracy in LVEF estimation~\cite{dai2022cyclical,ouyang2020video}.

Recent deep learning approaches have shown promise in automating LVEF prediction by leveraging large-scale echocardiographic video datasets~\cite{dai2021adaptive,ouyang2020video}. These models typically rely on supervised learning paradigms that require massive labeled videos, making them highly dependent on extensive manual annotations and creating significant bottlenecks for model scalability and adaptability across different clinical settings.
Furthermore, domain shifts due to variations in acquisition protocols and ultrasound manufacturers often degrade model performance when deployed in real-world clinical environments~\cite{chao2023comparative,dai2022cyclical,yan2020mri}. To address these challenges, it is crucial to develop data-efficient adaptation strategies that can generalize across diverse conditions with minimal labeled supervision~\cite{dai2023semi,li2024few}.

Vision-language models (VLMs), particularly CLIP-based architectures, have recently emerged as powerful tools in medical image analysis~\cite{chen2024uni,christensen2024vision,du2024teach}. By aligning visual and textual representations, CLIP enables models to learn rich semantic features from large-scale image-text pairs without requiring extensive task-specific annotations~\cite{radford2021learning}. EchoCLIP~\cite{christensen2024vision} represents the first attempt to apply CLIP to echocardiography, achieving promising results in image-text retrieval.
However, EchoCLIP~\cite{christensen2024vision} extracts a random frame from each video for image-text matching pretraining, and averages predictions across frames without modeling the temporal dynamics of the cardiac cycle. 
In addition, the diagnosis of various cardiac diseases and important indicators heavily rely on the accurate perception of the temporal changes in specific regions of the heart~\cite{ghorbani2020deep,hughes2021deep}. 
However, CLIP-based models are known to have limited fine-grained feature understanding, making them less effective in identifying subtle cardiac abnormalities.
Since LVEF prediction is inherently a video-based task, existing CLIP models fail to fully exploit the temporal and anatomical information embedded in echocardiographic sequences.

To overcome these limitations, we propose \textbf{CardiacCLIP}, a novel video-based CLIP adaptation designed for few-shot LVEF prediction from echocardiogram videos. 
Our method introduces two key components: MFL~(Multi Frame Learning), an attention-based frame aggregation module that selectively fuses frame-level information for temporal modeling; and EchoZoom, a multi-resolution input scaling strategy tailored for capturing fine-grained anatomical features from the apical four-chamber views.
MFL mitigates the redundancy of frame-wise features by learning an optimal weighting mechanism, while EchoZoom ensures that the model attends to diagnostically relevant cardiac regions by fusing multi-scale representations.
These two components enhance model robustness and generalization, allowing CardiacCLIP to achieve superior performance in data-limited settings.
Our contributions are summarized as follows:
\begin{enumerate}
    \item We introduce CardiacCLIP, a novel CLIP-based framework specifically designed for video-based echocardiography analysis, addressing the limitations of image-based CLIP models.
    \item We develop MFL, an attention-based frame fusion mechanism that effectively captures temporal dependencies in LVEF estimation.
    \item We propose EchoZoom, a multi-resolution scaling strategy that enhances the model’s ability to capture fine-grained structural details.
    \item We demonstrate that CardiacCLIP significantly outperforms existing methods in few-shot settings, achieving state-of-the-art performance.
\end{enumerate}

\section{CardiacCLIP for LVEF Prediction}

\subsection{Preliminary: Coarse-to-Fine Ordinal Regression}
LVEF estimation can be reformulated as an ordinal regression problem, where we first convert it as a classification task by discretizing the labels as different bins and treating each bin as an independent class, and then regress the specific values based on the classification results~\cite{du2024teach,rothe2015dex}.
The motivation for this is based on the fact that learning from a staged classification process is more effective and easier than directly learning from multiple precise values, especially in the imperfect data scenario~\cite{pintea2023step}.
This reformulation allows training with cross-entropy loss while maintaining numerical continuity via an MAE-based regression refinement~\cite{fu2018deep,wang2021uniformity}:
\begin{small}
\begin{align}
    L_{OR} = L_{CE} + L_{MAE}\quad.
\label{equ:ordinal regression}
\end{align}
\end{small}

For the coarse-to-fine framework, the coarse stage maps LVEF values into discrete bins, leveraging CLIP’s pretrained visual-textual alignment. This transforms the problem into a classification task, where text embeddings serve as classifier weights. The fine stage refines predictions via a lightweight MLP regressor, making the final estimation:
\begin{small}
    \begin{equation}
    y^{*}= \sum\limits_{i=1:k} p_i * \frac{b_i}{1+{\delta}_i} \quad,
\end{equation}
\end{small}

\noindent where $k$ is the number of classes, $p_i$ is the class probability, $b_i$ is the centre of $i_{th}$ mapped numerical group, and ${\delta}_i$ is the estimated shift from the regressor to make the bin interval learnable.
These two stages are trained end-to-end.

\subsection{Video-based LVEF Prediction} 
Leveraging its robust representation capabilities from pretraining on extensive image-text pairs, CLIP serves as a foundational model for various downstream tasks, including video recognition~\cite{rasheed2023fine}. 
Given an echocardiogram video $x_i \in \mathbb{R}^{T \times C \times H \times W}$ with $T$ frames and each frame is with the spatial dimension of $H \times W$, we process the video through the CLIP visual encoder $f_v(\cdot)$ to extract features: 
\begin{small}
    \begin{equation}
\label{clip visual}
    z^{v_i} = f_v(x_i)\quad,
\end{equation}
\end{small}
\noindent where the visual feature $z^{v_i} \in \mathbb{R}^{T \times C}$ and $C$ represents the feature dimension of the \texttt{[CLS]} token. 
Unlike previous CLIP methods for video adaptation that average frame-level features to obtain video representations~\cite{huang2024froster,rasheed2023fine}, we introduce MFL~(Multi Frame Learning), attention-based feature aggregation to capture critical cardiac dynamics.

For text features, we tokenize clinically relevant LVEF descriptions (e.g., \emph{``The left ventricular ejection fraction is estimated to be mildly reduced LVEF (45-54\%)''}) and embed them via the CLIP text encoder $f_t(\cdot)$:
\begin{small}
    \begin{equation} 
\label{clip text} 
    z^{t_j} = f_t(t_j)\quad, 
\end{equation} 
\end{small}
\noindent where $z^{t_j} \in \mathbb{R}^{C}$. 
To enrich text representations, we leverage GPT-4~\cite{achiam2023gpt} to generate diverse descriptions corresponding to LVEF intervals, enhancing data efficiency and serving as a form of text data augmentation during training.
Given the ground-truth category label $y_i$, the model is trained via a cross-entropy loss:
\begin{small}
\begin{align}
    L_{CE} = -\frac{1}{N}\sum_i^N\sum_j^K\hat{y}_{i,j}\log{y_{i,j}}\quad,
\label{eq: vido finetune}
\end{align}
\end{small}\noindent where $K$ denotes the total number of classes. 
Thus for video-based adaptation, the $L_{CE}$ in Loss~\ref{equ:ordinal regression} should be updated with Loss~\ref{eq: vido finetune}.

\vspace{-3mm}\begin{figure*}[ht]
\begin{center}
   \includegraphics[width=\textwidth]{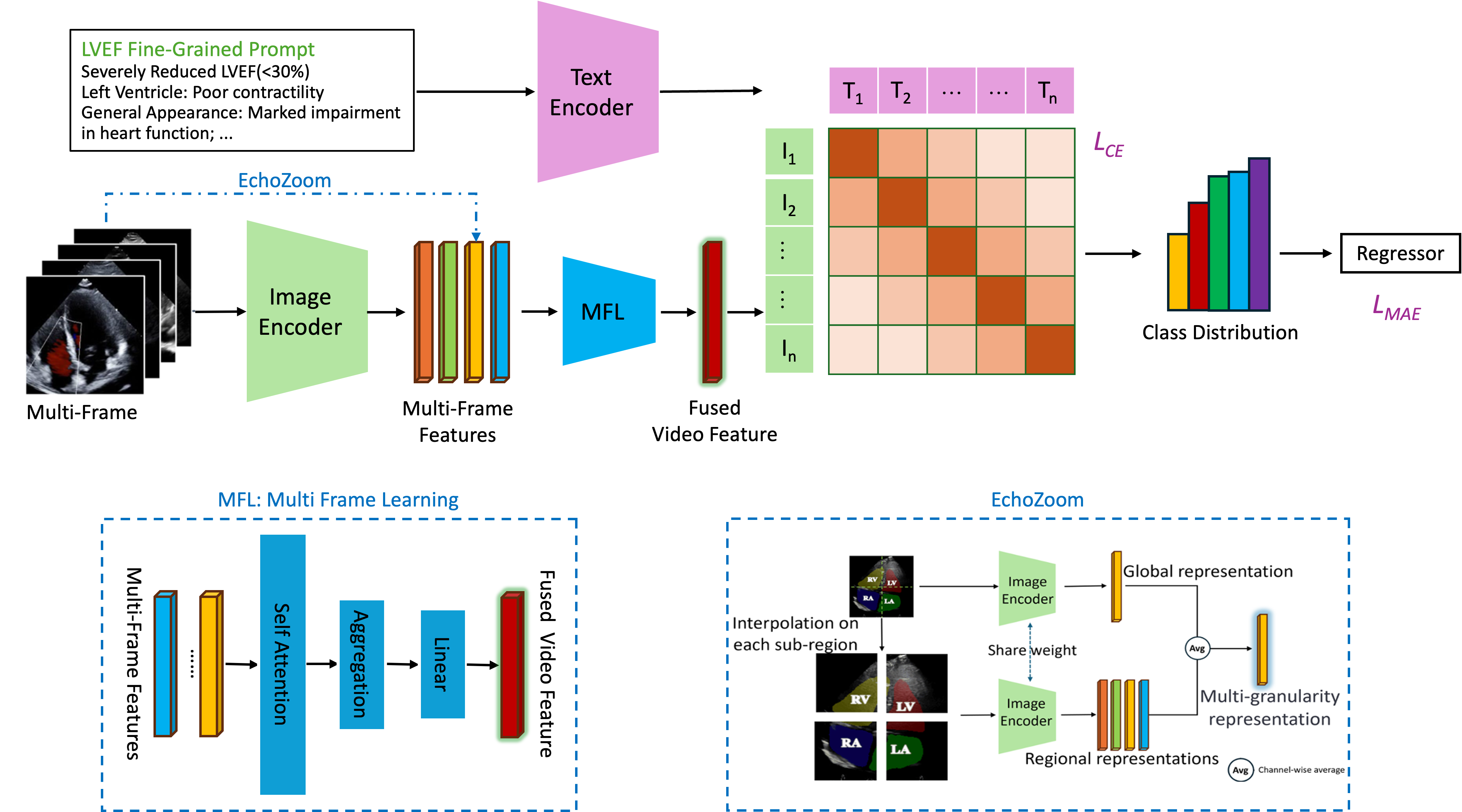}
\end{center}
\vspace{-4mm}
   \caption{\textbf{CardiacCLIP}: Video-based CLIP adaptation for few-shot LVEF prediction integrating multi frame learning and multi-scale representations.
   }
\label{fig:framework}
\end{figure*} \vspace{-8mm}

\subsection{Video Representation via Multi Frame Learning}

Typical CLIP-based video recognition models aggregate frame features via simple averaging~\cite{huang2024froster,rasheed2023fine}, overlooking temporal variability. In LVEF estimation, different frames contribute unequally due to varying cardiac contraction phases. Inspired by Multiple Instance Learning (MIL)~\cite{ilse2018attention,shao2021transmil,sudharshan2019multiple}, we introduce MFL, an attention-based fusion mechanism that prioritizes diagnostically relevant frames.

Given an input video sequence with \( B \) frames, we extract a set of frame-level features:
\begin{small}
    \begin{equation}
    F = [f_1, f_2, \dots, f_B] \in \mathbb{R}^{B \times C}\quad,
\end{equation}
\end{small}

\noindent where \( f_i \in \mathbb{R}^{C} \) represents the feature vector of the \( i \)-th frame, and \( C \) is the feature dimension. Instead of using average pooling, we introduce an attention mechanism to learn the relative importance of each frame dynamically.

\textbf{Frame Importance Estimation.}
we compute per-frame importance scores using a multi-layer attention network:

\begin{equation}
    s_i = W_3 \tanh(W_2 \tanh(W_1 f_i))\quad,
\end{equation}

\noindent where \( W_1, W_2, W_3 \) are learnable weight matrices.
These scores are normalized using softmax to ensure that the sum of the weights is equal to 1:

\begin{equation}
    \alpha_i = \frac{\exp(s_i)}{\sum_{j=1}^{B} \exp(s_j)}\quad,
\end{equation}

\noindent where \( \alpha_i \) represents the learned weight for frame \( i \), ensuring higher weights for informative frames while suppressing redundancy.

\textbf{Dynamic Feature Aggregation.}
The aggregated video representation is computed as:

\begin{equation}
    F_{\text{agg}} = \sum_{i=1}^{B} \alpha_i f_i\quad,
\end{equation}

\noindent where \( F_{\text{agg}} \in \mathbb{R}^{C} \) is the aggregated feature vector representing the entire video. This formulation ensures that the model prioritizes the most relevant frames.
Finally, the aggregated feature vector is passed through a linear projection layer:

\begin{equation}
    F_{\text{final}} = W_{\text{proj}} F_{\text{agg}}\quad,
\end{equation}

\noindent where \( W_{\text{proj}} \in \mathbb{R}^{C \times C} \) is a learnable projection matrix that refines the video representation.

By integrating MIL-inspired attention-based aggregation, our model learns to emphasize diagnostically-relevant frames, offering a more robust, adaptive, and interpretable approach to video-based LVEF prediction compared to the conventional average pooling strategy. 
In practice, we find that the input frame length plays a crucial role in the performance of feature aggregation and we discuss it in the ablation study.

\subsection{EchoZoom: Multi-Scale Cardiac Representation}
\label{EchoZoom}

Echocardiographic diagnosis relies on analyzing regional cardiac dynamics, particularly within the left and right ventricles and atria~\cite{dai2022cyclical,ghorbani2020deep}. Standard vision models process images at a fixed resolution, which limits the model’s ability to capture multi-scale anatomical variations.
EchoZoom enhances regional cardiac representation by applying multi-resolution input scaling. As shown in the lower right corner of Figure~\ref{fig:framework}, it processes images at multiple scales (e.g., $112^2$, $224^2$), enabling fine-grained structural analysis. 
Specifically, EchoZoom splits the $224^2$ image into four $112^2$ sub-images. These sub-images, along with the original $112^2$ image, are fed through the same pretrained model. The features extracted from these sub-images are then combined into a larger feature map corresponding to the $112^2$ image. This map is subsequently average-pooled to match the feature map size of the original $112^2$ image. The final output is the fused feature map generated across all scales.
This process enriches feature extraction without requiring additional parameters, reinforcing the model’s ability to recognize subtle morphological changes across varying resolutions.

\section{Experiments}

\subsection{Datasets and Experiment Settings}

\textbf{Dataset.} 
We evaluate our method on EchoNet-Dynamic~\cite{ouyang2019echonet}, a widely used benchmark dataset in echocardiography. It contains 10,036 apical four-chamber echocardiogram videos collected from Stanford University Hospital using five different ultrasound machines. Each video, averaging 175 frames, is resized to 112×112 and annotated with its corresponding LVEF label. The dataset is pre-split into 7,465 training, 1,288 validation, and 1,277 test samples.
For few-shot evaluation, we extract subsets from the training set following the 1/2/4/8-shot settings (Table~\ref{tab:few_shot}).

\noindent\textbf{Experiment Settings.}
Following EchoCLIP~\cite{christensen2024vision}, we adopt ConvNext-Base CLIP as the backbone for fair comparison, and the model’s pretraining dataset does not overlap with our current dataset.
Our model is optimized using RAdam~\cite{liu2019variance} for 100 epochs, starting with a learning rate of 5e-5, which is cosine-decayed to zero. 
Each input clip consists of 48 frames, sampled at a stride of 2, with a batch size of 2.
To construct a typical few-shot dataset, we discretize LVEF values into integer classes (1-100) and sample training examples accordingly, skipping any missing classes, as detailed in Table~\ref{tab:few_shot}.
We adopt Mean Absolute Error (MAE) and Root Mean Square Error (RMSE) as evaluation metrics to assess the model performance.

\vspace{-3mm}\begin{table*}[h]
\centering
\caption{Sample counts under few-shot settings for EchoNet-Dynamic dataset.}
\scalebox{0.9}{
\setlength{\tabcolsep}{5pt} 
\begin{tabular}{c|cccc}
    \toprule
    Dataset & 1-shot & 2-shot & 4-shot & 8-shot \\
    \midrule
    EchoNet-Dynamic& 84 & 162 & 307 & 570 \\
    \bottomrule
\end{tabular}
}
\label{tab:few_shot}
\end{table*}\vspace{-8mm}

\subsection{Results under Few-shot Setting}
We mainly compare CardiacCLIP against two categories of methods:
1) Traditional LVEF prediction methods (video-based models trained end-to-end);
2) CLIP-based methods (pretrained VLMs adapted to echocardiography).

\noindent Table~\ref{tab:EchoNet-Dynamic} presents the results. CardiacCLIP consistently outperforms existing methods, achieving a 2.07 MAE reduction over EchoNet~\cite{ouyang2020video} in the 1-shot setting. Similar performance gains can be observed across other shot settings, highlighting the effectiveness of our method.
The performance improvement diminishes as training data increases, a typical phenomenon in few-shot learning.

\begin{table*}[ht]
  \centering
  \tabcolsep=0.8pt 
    \caption{Comparison of different SOTA methods on EchoNet-Dynamic dataset under few-shot setting.
    \textcolor{red}{\footnotesize(Num)} indicates the performance improvement compared to EchoNet~\cite{ouyang2020video}.
    }
    \scalebox{0.95}{
    \begin{tabular}{l@{\hspace{2mm}}cccc@{\hspace{2mm}}|@{\hspace{2mm}}cccc}
    \toprule
    \multirow{2}{*}{\textbf{Method}} & \multicolumn{4}{c}{\textbf{MAE $\downarrow$}} & \multicolumn{4}{c}{\textbf{RMSE $\downarrow$}} \\ 
    \cmidrule{2-5} \cmidrule{6-9}
    & 1-shot & 2-shot & 4-shot & 8-shot & 1-shot & 2-shot & 4-shot & 8-shot \\ 
    \midrule
    \textbf{\footnotesize \emph{Traditional}} \\
    EchoNet~\cite{ouyang2020video} & 9.32 & 9.17 & 7.49 & 6.81 & 12.11 & 11.89 & 9.92 & 9.19 \\
    AdaCon~\cite{dai2021adaptive} & 9.52 & 8.83 & 7.35 & 7.02 & 12.17 & 11.56 & 9.91 & 9.46 \\
    C-Mixup~\cite{yao2022c} & 9.23 & 9.22 & 9.13 & 7.59 & 12.23 & 12.14 & 11.87 & 9.77 \\
    BalancedMSE~\cite{ren2022balanced}& 8.50 & 7.87 & 7.55 & 7.05 & 11.22 & 10.18 & 9.65 & 9.35 \\
    \midrule
    \textbf{\footnotesize \emph{CLIP-based}} \\
    EchoCLIP~\cite{christensen2024vision} & 10.54 & 9.74 & 9.22 & 9.12 & 13.18 & 12.18 & 11.53 & 11.40 \\
    NumCLIP~\cite{du2024teach} & 7.91 & 7.56 & 7.68 & 6.96 & 9.89 & 9.45 & 9.60 & 8.70 \\
    \midrule
    \textbf{\footnotesize \emph{Our Method}} \\
    CardiacCLIP & 7.25 & 7.11 & 6.79 & 6.42 & 9.06 & 8.89 & 8.49 & 8.02 \\
    {{\color[HTML]{CB0000}$\Delta$}} & \textcolor{red}{\scriptsize(2.07)} & \textcolor{red}{\scriptsize(2.06)} & \textcolor{red}{\scriptsize(0.70)} & \textcolor{red}{\scriptsize(0.39)} & \textcolor{red}{\scriptsize(3.05)} & \textcolor{red}{\scriptsize(3.00)} & \textcolor{red}{\scriptsize(1.43)} & \textcolor{red}{\scriptsize(1.17)} \\
    \bottomrule
    \end{tabular}%
    }
    \label{tab:EchoNet-Dynamic}
\end{table*}

\subsection{Ablation Study}

We conduct detailed ablation experiments to analyze the contributions of model components, input frame length, loss functions, and aggregation methods, under 1-shot setting.

\noindent \textbf{Effect of Model Components.}
Table~\ref{table:combined-ablation} shows the impact of EchoZoom and MFL,
demonstrating that both modules contribute to enhanced model performance, with their joint combination achieving the best result.


\begin{table}[h]
    \centering
    \small
    \caption{Ablation study of CardiacCLIP on EchoNet-Dynamic dataset.}
    \label{table:combined-ablation} 
    \setlength{\tabcolsep}{5pt} 
    \scalebox{0.90}{
    \begin{tabular}{l|c|c|c}
        \toprule
        \textbf{Ablation Study} & \textbf{EchoZoom} & \textbf{MFL} & \textbf{MAE $\downarrow$} \\
        \midrule
        Base & \ding{55} & \ding{55} & 7.91 \\
        w/o EchoZoom & \ding{55} & \ding{51} & 7.42 \\   
        w/o MFL & \ding{51} & \ding{55} & 7.50 \\
        Ours & \ding{51}& \ding{51} & \textbf{7.25} \\
        \bottomrule
    \end{tabular}
    }
\end{table}

\noindent \textbf{Effect of Frame Length.}
Table~\ref{tab:frame_length} presents the impact of input frame length on model performance. While shorter frame length (e.g., 16 frames) result in higher MAE (7.89), increasing the frame length initially improves performance, with the best MAE achieved at 48 frames (7.25). Beyond this, performance fluctuates slightly, indicating that longer sequences do not necessarily enhance feature extraction, likely due to increased redundancy in the input.

\noindent\textbf{Effect of MFL Modules.}
Table~\ref{tab:attention_strategy} examines various design choices within MFL. Our proposed MFL achieves an MAE of 7.25, while removing the final projector increases the error to 7.53, highlighting the importance of feature transformation. Introducing a nonlinear projector yields a slight performance drop to 7.45, likely because the aggregated features are already well-structured. Incorporating a gated recurrent unit (GRU) further degrades performance, increasing the MAE to 8.26, suggesting excessive temporal dependencies may lead to overfitting.

\begin{table*}[ht] \vspace{-4mm}
\centering
\begin{minipage}{0.48\linewidth}
    \centering
    \caption{Ablation on frame length.}
    \label{tab:frame_length}
    \setlength{\tabcolsep}{3pt} 
    \renewcommand{\arraystretch}{1.2} 
    \scalebox{0.9}{
    \begin{tabular}{c|c}
    \toprule
    \textbf{Frame Length} & \textbf{MAE $\downarrow$} \\ 
    \hline
    16  &  7.89 \\
    36  & 7.64 \\ 
    \textbf{48}  & \textbf{7.25} \\ 
    54  & 7.72 \\ 
    64  & 7.38 \\ 
    96  & 7.92 \\ 
    128 & 7.94 \\ 
    \bottomrule
    \end{tabular}
    }
    
\end{minipage}
\hfill
\begin{minipage}{0.5\linewidth}
    \centering
    \caption{Ablation on MFL modules.}
    \label{tab:attention_strategy}
    \setlength{\tabcolsep}{3pt} 
    \renewcommand{\arraystretch}{1.2} 
    \scalebox{0.95}{
    \begin{tabular}{l|c}
    \toprule
    \textbf{Aggregation} & \textbf{MAE $\downarrow$} \\ 
    \hline
    \textbf{MFL}  & \textbf{7.25} \\ 
    w/o Projector  & 7.53 \\ 
    w/ Nonlinear Projector  & 7.45 \\ 
    w/ GRU  & 8.26 \\ 
    \bottomrule
    \end{tabular}
    }
    
\end{minipage}
\end{table*}

\begin{table*}[ht]   \vspace{-9mm}
\centering
\begin{minipage}{0.48\linewidth}
    \centering
    \caption{Ablation on regression loss.}
    \label{tab:loss}
    \setlength{\tabcolsep}{3pt} 
    \renewcommand{\arraystretch}{1.2} 
    \scalebox{0.95}{
    \begin{tabular}{l|c}
    \toprule
    \textbf{Regression Loss} & \textbf{MAE $\downarrow$} \\ 
    \hline
    \textbf{MAE}      & \textbf{7.25} \\ 
    SmoothL1  & 7.36 \\ 
    Huber~\cite{huber1992robust}   & 7.57 \\ 
    MSE       & 7.75 \\ 
    \bottomrule
    \end{tabular}
    }
\end{minipage}
\hfill
\begin{minipage}{0.5\linewidth}
    \centering
    \caption{Ablation on aggregation methods.}
    \label{tab:aggregation_methods}
    \setlength{\tabcolsep}{3pt} 
    \renewcommand{\arraystretch}{1.2} 
    \scalebox{0.95}{
    \begin{tabular}{l|c}
    \toprule
    \textbf{Aggregation} & \textbf{MAE $\downarrow$} \\ 
    \hline
    \textbf{MFL}  & \textbf{7.25} \\ 
    Multi-Head & 10.47 \\ 
    Multi-Head+GRU  & 8.3 \\ 
    \bottomrule
    \end{tabular}
    }
   
\end{minipage}
\end{table*}

\noindent \textbf{Effect of Regression Loss.}
Table~\ref{tab:loss} investigates how different regression loss functions impact model performance. The standard MAE loss achieves the lowest error (7.25), while SmoothL1 (7.36) and Huber (7.57) introduce slight performance degradation. MSE loss performs the worst (7.75), likely due to its sensitivity to large errors, which may disproportionately penalize outliers.

\noindent \textbf{Effect of Aggregation Methods.}
Table~\ref{tab:aggregation_methods} evaluates the impact of different aggregation methods. Our MFL achieves the best performance, whereas Multi-Head Attention significantly degrades accuracy, increasing the MAE to 10.47, likely due to feature distortion caused by excessive attention heads. In video recognition tasks like LVEF prediction, only a subset of key frames holds critical diagnostic information. Introducing a GRU into the Multi-Head approach improves performance to 8.3, suggesting that temporal modeling can partially counteract attention-related issues. These findings are consistent with the observations in Table~\ref{tab:attention_strategy}.

\section{Conclusion}
In this paper, we introduce \textbf{CardiacCLIP}, a novel framework for LVEF estimation from echocardiogram videos, extending CLIP-based models to effectively capture both spatial and temporal cardiac features. Our method addresses the limitations of prior approaches by incorporating Multi-frame Learning (MFL) for adaptive temporal feature aggregation and EchoZoom, a multi-scale input strategy that enhances the representation of key anatomical structures. Through a few-shot learning paradigm, CardiacCLIP demonstrates strong generalization with limited labeled data, making it well-suited for clinical applications. Extensive experiments on the EchoNet-Dynamic dataset validate the effectiveness of our method, achieving state-of-the-art performance in few-shot settings. These results highlight the potential of CardiacCLIP as a robust and data-efficient solution for automated echocardiographic analysis, paving the way for improved cardiac disease diagnosis in real-world scenarios.



    

\begin{credits}
\subsubsection{\ackname} 
This work was supported by a research grant from the Joint Research Scheme (JRS) under The National Natural Science Foundation of China (NSFC) and the Research Grants Council (RGC) of Hong Kong (Project No. N\_HKUST654/24) as well as a grant from the National Natural Science Foundation of China (Grant No. 62306254).

\subsubsection{\discintname}
The authors have no competing interests to declare that are relevant to the content of this article.
\end{credits}

%
%
%
\bibliographystyle{splncs04}
\bibliography{Paper-0034}
%

\end{document}